\documentclass[runningheads]{llncs}
\usepackage[T1]{fontenc}
\usepackage{graphicx,verbatim}
\usepackage{amssymb}
\usepackage{hyperref}
\begin{document}
\title{Query Nearby: Offset-Adjusted Mask2Former enhances small-organ segmentation}
%

\author{Xin Zhang\inst{1} \and Dongdong Meng\inst{2} \and Sheng Li\inst{3}}  
\authorrunning{Zhang et al.}
\institute{School of Public health, Peking University, Beijing, 100871, China \\
    \email{2110306225@stu.pku.edu.cn} \\ 
    \and School of Physics, Peking University, Beijing, 100871, China \\
    \email{dongdongmeng@pku.edu.cn} \\
    \and School of Computer Science, Peking University, Beijing, 100871, China \\
    \email{lisheng@pku.edu.cn}}

\maketitle              
\begin{abstract}
Medical segmentation plays an important role in clinical applications like radiation therapy and surgical guidance, but acquiring clinically acceptable results is difficult. In recent years, progress has been witnessed with the success of utilizing transformer-like models, such as combining the attention mechanism with CNN. In particular, transformer-based segmentation models can extract global information more effectively, compensating for the drawbacks of CNN modules that focus on local features. However, utilizing transformer architecture is not easy, because training transformer-based models can be resource-demanding. Moreover, due to the distinct characteristics in the medical field, especially when encountering mid-sized and small organs with compact regions, their results often seem unsatisfactory. For example, using ViT to segment medical images directly only gives a DSC of less than 50\%, which is far lower than the clinically acceptable score of 80\%.
In this paper, we used Mask2Former with deformable attention to reduce computation and proposed offset adjustment strategies to encourage sampling points within the same organs during attention weights computation, thereby integrating compact foreground information better. Additionally, we utilized the 4th feature map in Mask2Former to provide a coarse location of organs, and employed an FCN-based auxiliary head to help train Mask2Former more quickly using Dice loss. We show that our model achieves SOTA (State-of-the-Art) performance on the HaNSeg and SegRap2023 datasets, especially on mid-sized and small organs.Our code is available at link \href{https://github.com/earis/Offsetadjustment\_Background-location\_Decoder\_Mask2former}.


\end{abstract}
\section{Introduction}

Medical segmentation plays an important role in clinical applications, including surgical guidance and radiation therapy \cite{RN1,Lei2024ConDSegAG}. To achieve clinically acceptable results, segmentation must consider the problem of how to combine local information and global information; Convolutional Neural Networks (CNNs) can integrate local information, but less effective on global information compared to transformer architectures \cite{RN13}, but how to reduce the heavy computational burden and data demand is important. Deformable DETR\cite{RN31}, featuring its deformable attention computation aimed at reducing the burden of training by querying sampled points instead of all pixels, provides insights for leveraing transformers. But utitlizing them directly is not enough. As depicted in Fig. 1, using transformer-based models directly would be accuracy-limited, especially for small organs as depicted in Fig. 2, which appear less frequently. Therefore, to utilize mask-attention-transformer-based models for clinical usage, enhancing the segmentation accuracy of small organs is vital to achieving better medical segmentations.

 In this paper, we used a deformable-attention-based transformer, Mask2Former, directly as our framework. Based on the sampling policy in attention computing to get points to be queried (computing offsets to the querying point and adding them onto the coordinates to get sampled points), we complicated the process to better segment mid-sized and small organs. In fact, we encouraged points to query nearby points that are suspected to be within the same organ, since the regions of the foreground are often small and compact. Based on this mechanism, we contrasted three strategies of offset adjustment to help enhance accuracy on relatively small organs. This process is integrated into any transformer that uses Deformable attention to compute attention weights, allowing any lightweight transformer to apply it. Besides, we merged the unused largest feature maps into the training process and added an FCN-based branch to help obtain coarse localization information and enhance contrast to the background.
Our contributions can be concluded into three parts as follows:
\begin{enumerate}

    \item We modified Mask2Former with three strategies to control offset computing, thereby enhancing the model's general ability for small object segmentation. This can be applied to any lightweight transformer that utilizes deformable attention.
    \item We adopted a feature fusion strategy to utilize the full information extracted inside Mask2Former itself. Since these feature maps already exist, we do not add the burden of extracting more features, which better combines global information for segmentation. 
    \item We added an FCN-based branch to help obtain coarse localization information and enhance contrast to the background, thus helping the model quickly find the location of organs and focus on better margin improvement. This alleviates the problem of vast background disturbance in typical medical screens.

\end{enumerate}
 \subsection{Related work}
 In past years, methods based on UNet \cite{RN2} have achieved significant progress with the advantage of CNNs. It can integrate local information, but less effective on global information compared to transformer architectures\cite{RN13}. However, due to the heavy computational burden and data demand, transformer-relevant methods \cite{RN16,RN18} are often adopted by adding attention mechanism alone, such as TransUnet\cite{RN16}, UNETR \cite{RN17} and CSAF-CNN \cite{RN18}. Transformer architecture as based model is not as popular. For example, in the SegRap2023 challenge, to segment head and neck organs, the top-10 methods are all based on nnUNet\cite{RN15}. However, DETR\cite{RN98} and Deformable DETR\cite{RN31} inspire researchers to use lightweight deformable attention to reduce resource cost while keeping accuracy, and many trials are based on them, like SAM-DETR\cite{Zhang2022AcceleratingDC}, MaskFormer\cite{RN3}, and Mask2Former. There have been many trials of adopting Mask2Former to segment medical images \cite{RN22,RN23}, and enhancing accuracy with a CNN auxiliary branch is effective even without a complicated architecture. Besides, Takikawa \cite{RN27} proposed Gated-SCNN to add an auxiliary branch to learn the marginal information of objects.

\begin{figure}
\includegraphics[width=\textwidth]{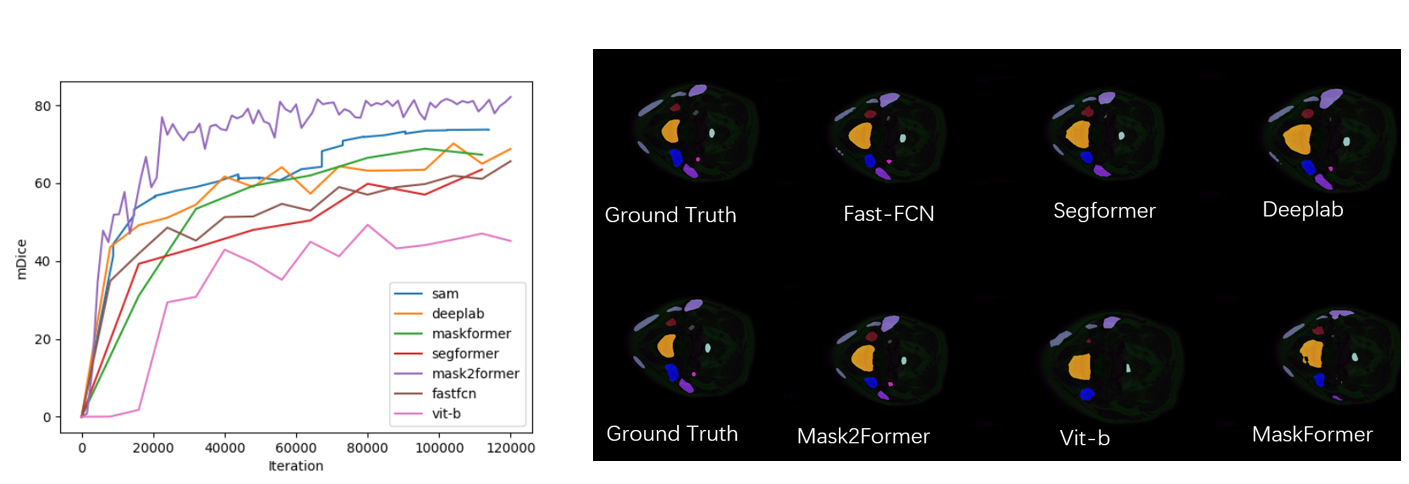}
\caption{Results on Different Models for the HaN-Seg Dataset and Visualization of Selected Results.} \label{fig1}
\end{figure}

\section{Methods}
\subsection{Overview}
An overview of our framework is presented in Fig. 2. We first use a standard backbone, such as ResNet-50, to extract feature maps and use them as inputs to the decoder as well as one FCN auxiliary head. Then, during the computation of attention weights, we apply adjustment strategies when acquiring the offset so that it is more efficient on mid-sized and small organs. The fusion strategy is performed in the next process. In Fig. 2, only one fusion strategy adopted in our model, the late fusion, is depicted. During the attention interaction with feature maps of four sizes, the fourth one, which is the biggest, is used for computing coarse embeddings (i.e., memories) and the weights of the attention layer. Then, other feature maps containing more local information are utilized to obtain better memories. The two are integrated with a residual connection. Finally, with outputs from the decoder, we use another auxiliary head for quick coarse location segmentation against the background to help accelerate training. The mask and class losses are then computed for loss backward. The details of the three kinds of fusion strategies, offset adjustment strategies, and the auxiliary branch will be discussed in the following sections.

\begin{figure}
\includegraphics[width=0.8\textwidth]{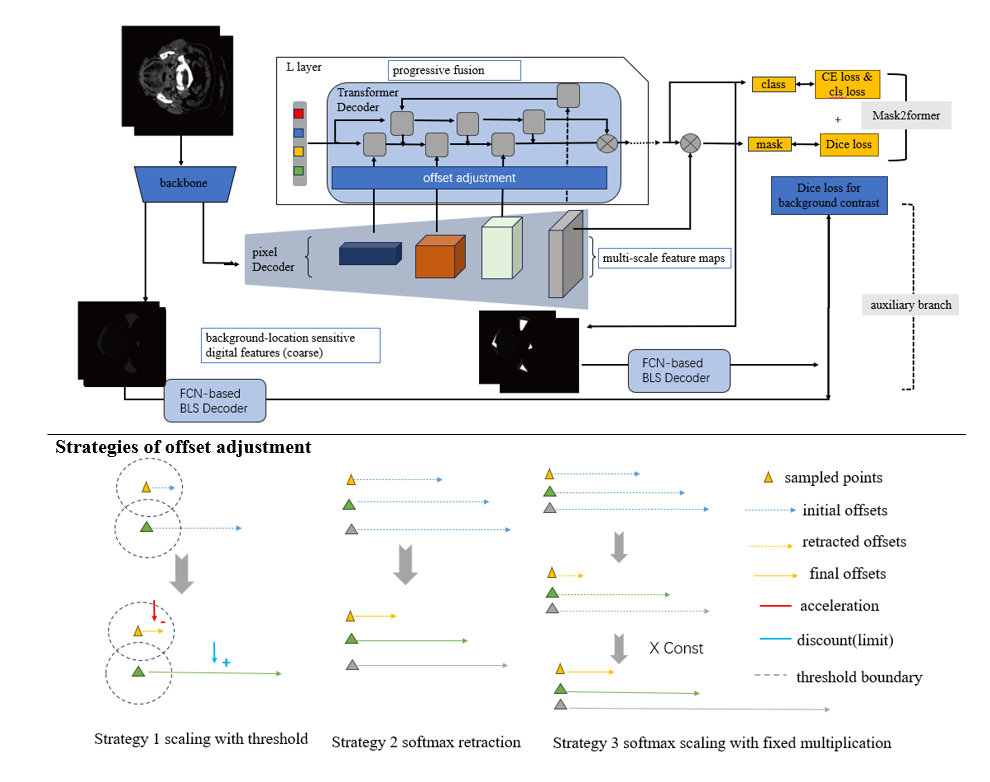}
\centering
\caption{Overview of Our Offset-Adjusted Mask2Former with Feature Fusion, Background-Location Sensitive Auxiliary Branch, and Three Offset Adjustment Strategies.} \label{fig2}
\end{figure}

\subsection{Offset Adjustment Methods for Middle and Small Organ Segmentation Improvement}
As mentioned above, Mask2Former \cite{RN3} utilizes a simple linear layer to compute the offset for each input point. The offset is used to acquire candidate points to be queried by the original point, making the attention weights more efficient. However in medical images, the background often occupies a large proportion, while target organs typically occupy relatively small regions. Therefore, sampling reference points uniformly across the entire image is imbalanced and unnecessary. It is expected that Mask2Former can sense that small organs commonly appear in compact regions by not generating excessively large offsets for reference points.

To address this, a simple offset adjustment strategy was added, which first deepens the offset generation layers and then controls the offsets to be larger or smaller. For example, with a fixed threshold, offsets larger than it are discouraged by dividing them by a certain constant, to promote a focus on local information. This is strategy 1, a manual approach by setting the threshold. Strategy 2 uses the softmax function alone to retract offsets for small objects. Strategy 3, based on the former, uses the softmax function first to establish relationships between offsets at different points and then enlarges this gap by applying a scaling constant. This encourages the sampling trend a point is likely to develop. This pixel-wise algebraic operation aims not only to address the issue of small object segmentation but also to improve the efficiency of the attention acquisition process, which is why transformer slows down training in the classic paradigm.
We conducted various experiments to demonstrate the effectiveness of our proposals, and the results imply that the Softmax method, when multiplied by a constant greater than 1, is particularly outstanding.

\subsection{Feature fusion with unused maps}
In the naïve Mask2Former model, multi-scale features are used for attention computation in the encoder. The process involves concatenating three different levels of feature maps and extending the 2D maps into a one-dimensional feature vectors. However, the fusion operation of stacking is direct but indifferent to the relationship between different feature maps. Meanwhile, the fourth feature map, which has a channel size of 256 and contains low-resolution information, is not adopted in the original model. We aim to use this extra information for global guidance by progressively merging it into the encoder layers. Without burden the feature extraction, This approach provides auxiliary organ location information and margin information for the attention query processing. To be comprehensive, we modified Mask2Former to incorporate unused fourth feature maps at different positions, such as before the querying process , after memory vectors are computed, and during the computation; the fusion after the memory vectors is most effective.

\subsection{The background-location sensitive auxiliary decoder}
In medical images, the background, where many pixels are zero, often distracts the attention module from finding what it truly cares about. Thus, adding such an auxiliary head can accelerate training. We do not want the transformer-based encoder to get lost in the vast black areas of medical images. Adding CNN auxiliary heads is useful, and we are interested with more complex designs. Therefore, we adopted a background-location sensitive auxiliary branch, whose basic architecture is FCN—a pioneer and practical network—to extract the basic location information of each organ and the contrastive background information for margin refinement. This decoder was trained on a lightweight class segmentation task to quickly obtain the shape of organs against the background contrast. This module is independent of the inherent feature fusion inside Mask2Former, and thus we treat it as novel information that also plays the role of coarse localization, which can be used to any model to refine their margins.

\section{Experiment}
\subsection{Datasets}
The HaN-Seg dataset \cite{RN35} consists of 56 cases of CT and MRI modalities, but only 42 are provided open. In our method, we first selected 33 out of the 42 cases with only CT images to compare the Dice score against the baseline. Then, using all 42 cases with after registration of CT and MRI as described in \cite{RN36}, we tested our best results. The SegRap dataset \cite{RN15} contains 120 training cases and 20 test cases of CT and CT contrast volumes (not released). We trained them with the same configurations as the former dataset. Interestingly, we found that a simple trick in data preprocessin——using the original image stacked with two that were enlarged to 2× and then reduced to 0.5×, forming a three-channel image——could enhance performance without any additional complexity, so we adopted this. The CT contrast modality was not used. The training, validation, and test datasets contain 100, 7, and 10 images, respectively. We used a batch size of 2 for training, with the Adam optimizer and a loss function combining Dice loss and cross-entropy loss. The models were trained on GPU RTX 4090 × 8. The process was implemented using MMLab \cite{RN37}.

\subsection{Comparison to State-of-the-Art Methods}

On Table 1, compared to a 2D baseline model with the same settings, our improved model achieved a 10.83\% increase in mDice, and even outperformed some 3D models trained with two modalities. With MRI added, we achieved results that surpass the current SOTA on this dataset with SegReg \cite{RN36}, which is an nnUNet model assisted by non-linear registration to match CT and MRI information, by 1.4\% in mIoU and 0.35\% in mDice. On the SegRap2023 Dataset, using the Mask2Fomer assisted with preprocessing trick, the auxiliary branch, and offset adjustment, we achieved comparable results with an average Dice score 1.17\% better than the SOTA, while the top ten teams in \cite{RN15} all used nnUNet architecture.

In Fig. 3, we can find that the results of Cochlea and OpticNerve are improved dramatically. Mask2Former also outperforms in mid-sized organs such as the Mandible Bone, Spinal Cord, and Oral Cavity. There is generally an improvement, except for unstable results in symmetric or irregular organs like the Lacrimal Gland. 
As for application, previous work \cite{RN1,RN39,RN40} has performed assessments and found that most clinically acceptable results have a Dice score higher than 80\%. For the average Dice score, both HaNSeg and SegRap2023 are acceptable, but there are still some organs with insufficient results, such as the Lacrimal and Hippocampus, which are also mentioned by the supplier \cite{RN15}.  

\begin{table}
\caption{Our results compared to other models.}\label{tab1}
\begin{tabular}{|c|c|c|c|c|}
\hline
\textbf{model} & \textbf{Data used} & \textbf{Modality} & \textbf{mDice} & \textbf{mIoU } \\ \hline
        {\bfseries HaN-Seg} & ~ & ~ & ~ & ~ \\ 
        nnU-Net & 35 cases & CT & 58.69 & -  \\ 
        Ours & 35 cases  & CT & 72.26 (val) & 54.19 (val)  \\ 
        nnU-Net & 35 cases& CT/MR\_TI & 61.43 & -  \\ 
        nnU-Net(baseline) \cite{RN36}  & 56 cases & CT/MR\_T1 & 64.48 & 50.88  \\ 
        SegReg \cite{RN36} & 56 cases & MR\_T1 & 68.03 & 53.39  \\ 
        SegReg(SOTA)\cite{RN36} & 56 cases & CT/MR\_T1 & 81.26 & 69.05  \\ 
       
        Ours & 42 cases & CT/MR\_T1 & {\bfseries 81.60} & {\bfseries70.44}  \\ \hline
        {\bfseries SegRap2023} & ~ & ~ & ~ & ~ \\ 
        nnU-Net (baseline)\cite{RN15} & 140 cases& CT/ CT contrast & 84.65 & -  \\  
        Y. Zhong et al.\cite{RN15}(SOTA)& 140 cases & CT/ CT contrast & 86.70 & -  \\ 
        MaskFormer\cite{RN4} & 120 cases & CT only & 58.87 & -  \\ 
        FastFCN\cite{Wu2019FastFCNRD} & 120 cases & CT only & 61.99 & -  \\ 
        SegFormer\cite{RN14} & 120 cases & CT only & 69.87 & -  \\ 
        Deeplab\cite{RN12} & 120 cases & CT only & 70.45 & -  \\ 
        
        Naive Mask2Former& 120 cases & CT only & 84.18 & -  \\ 
        Ours*  & 120 cases & CT only & {\bfseries87.77} & -  \\ \hline
        \multicolumn{5}{p{11cm}}{{*In SegRap, some organs are overlapped; we just count those not overlapped regions of different organs: the overlapped regions are never big, and sometimes even negligible\cite{RN15} }} \\ \hline
\end{tabular}
\end{table}

\begin{figure}
\centering
\includegraphics[width=0.9\textwidth]{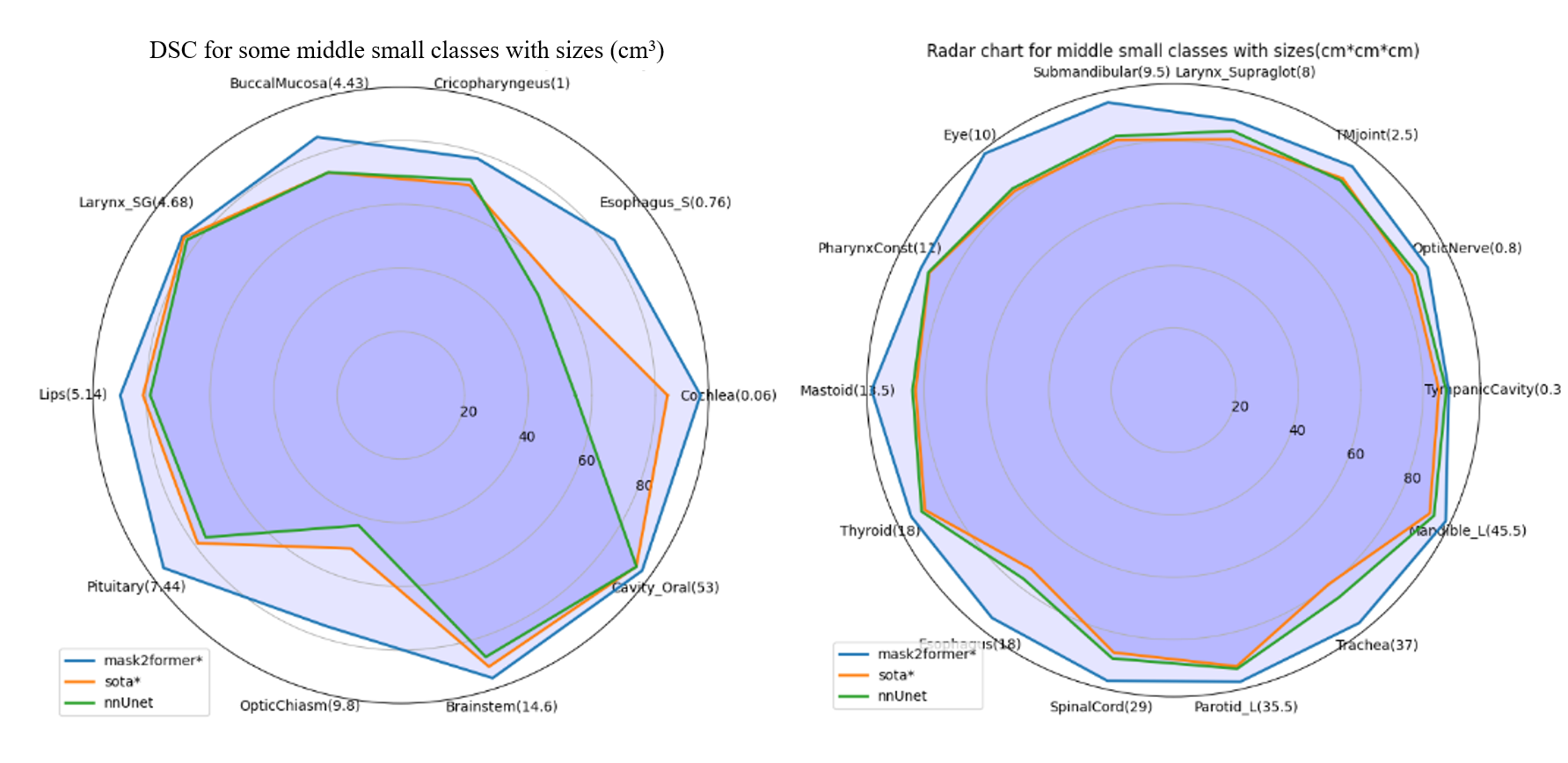}
\caption{Comparison in some mid-sized and small organ classes with nn-UNet baseline and previous SOTA.} \label{fig3}
\end{figure}

\begin{figure}
\includegraphics[width=\textwidth]{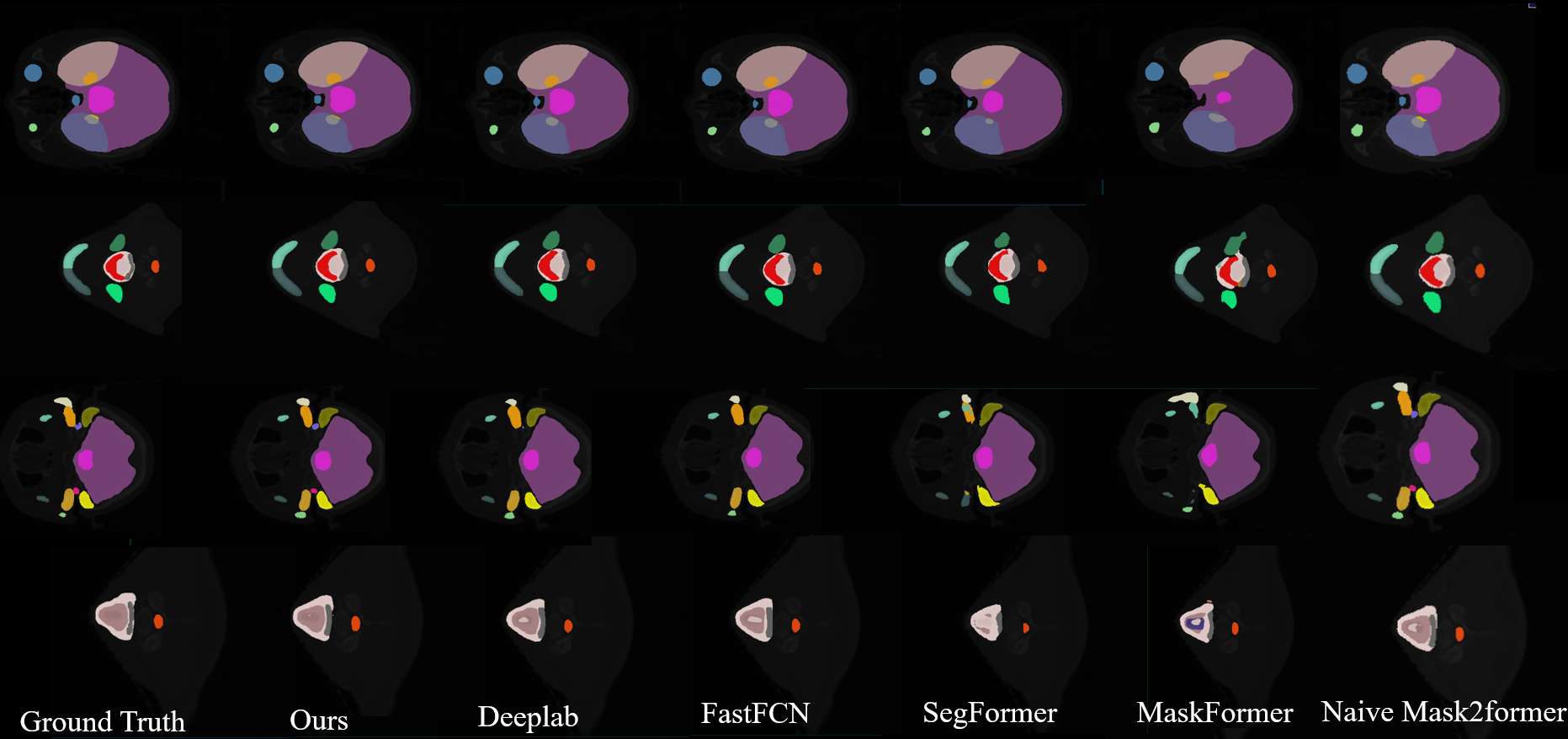}
\caption{Qualitative results of some re-implemented models on SegRap2023.} \label{fig4}
\end{figure}

\section{Ablation experiments}

We tested our models using naïve Mask2Former versus with additional modules and strategies on the SegRap2023 Dataset. 
On SegRap2023 dataset, compared to the naïve version, our offset adjustment strategy, feature fusion, and BLS Decoder all help improve the results. The data perprocessing trick and offset adjustment strategy with BLS helps to increase the Acc by 1.8\%, mDice by 2\% Adding more BLS increases the DSC to 87.7\% and the Acc to 84.13\%, representing an increase of 3\% to 3.5\% over the naïve Mask2Former. However, combining fusion strategy together helps the model gain improvements of approximately 3.2\% in mDice, and 2.2\% in mAcc compared to original version but not better than without the feature fusion, and we attribute it to the overfitting.

\begin{table}
\caption{Ablation test on SegRap2023. Sigmoid*2: offset adjustment strategies 3 with constant 2; FF: feature fusion; BLS: Background-Location sensitive decoder head; BLS(2): using 2 auxiliary heads}\label{tab1}
\centering
\begin{tabular}{|c|c|c|}

\hline
Configs  & mDice & mAcc \\ \hline
        Naive & 84.19 & 82.01  \\ \
        Dataset preprocessing trick   & 87.10  &83.16\\ 
        trick+OA (Strategy 1) &87.24 & 83.73\\
        trick+FF (inside) &88.01& 85.16\\
        trick+FF (late) &87.82	& 85.01\\
        trick+Sigmoid*2+BLS   & 87.28 & 83.87 \\ 
        trick+Sigmoid*2+BLS(2)  & 87.77 & 84.13 \\ 
        trick+Sigmoid*2+FF+BLS(2)   & 87.47 & 84.21 \\ \hline

\hline
\end{tabular}
\end{table}

\section{Conclusion}

We proposed an improved model of Mask2Former and tested its accuracy on the HaN-Seg dataset using incomplete information from 2D slices. The offset adjustment mechanism, along with the feature fusion mechanism in the encoder, prove to enhance the model's performance in segmenting organs of different sizes. Additionally, the background-location sensitive decoder helps improve coarse-to-fine segmentation. All these improvements contribute to our model's capacity to segment multi-organs precisely, even outperforming the baseline or SOTA models that are trained with full modality and 3D spatial information.

%
%
%
\bibliographystyle{splncs04}
\bibliography{ref}

\begin{thebibliography}{10}
\providecommand{\url}[1]{\texttt{#1}}
\providecommand{\urlprefix}{URL }
\providecommand{\doi}[1]{https://doi.org/#1}

\bibitem{RN98}
Carion, N., Massa, F., Synnaeve, G., Usunier, N., Kirillov, A., Zagoruyko, S.: End-to-end object detection with transformers. ArXiv  \textbf{abs/2005.12872} (2020)

\bibitem{RN16}
Chen, J., Lu, Y., Yu, Q., Luo, X., Adeli, E., Wang, Y., Lu, L., Yuille, A., Zhou, Y.: TransUNet: Transformers Make Strong Encoders for Medical Image Segmentation (2021). \doi{10.48550/arXiv.2102.04306}

\bibitem{RN12}
Chen, L.C., Papandreou, G., Kokkinos, I., Murphy, K., Yuille, A.L.: Deeplab: Semantic image segmentation with deep convolutional nets, atrous convolution, and fully connected crfs. IEEE Transactions on Pattern Analysis and Machine Intelligence  \textbf{40}(4),  834--848 (2018). \doi{10.1109/TPAMI.2017.2699184}

\bibitem{RN1}
Chen, X., Sun, S., Bai, N., Han, K., Liu, Q., Yao, S., Tang, H., Zhang, C., Lu, Z., Huang, Q., Zhao, G., Xu, Y., Chen, T., Xie, X., Liu, Y.: A deep learning-based auto-segmentation system for organs-at-risk on whole-body computed tomography images for radiation therapy. Radiotherapy and Oncology  \textbf{160},  175--184 (2021). \doi{https://doi.org/10.1016/j.radonc.2021.04.019}, \url{https://www.sciencedirect.com/science/article/pii/S0167814021062174}

\bibitem{RN3}
Cheng, B., Misra, I., Schwing, A.G., Kirillov, A., Girdhar, R.: Masked-attention mask transformer for universal image segmentation. In: 2022 IEEE/CVF Conference on Computer Vision and Pattern Recognition (CVPR). pp. 1280--1289. \doi{10.1109/CVPR52688.2022.00135}

\bibitem{RN4}
Cheng, B., Schwing, A.G., Kirillov, A.: Per-pixel classification is not all you need for semantic segmentation. In: Neural Information Processing Systems

\bibitem{RN37}
Contributors, M.: {MMSegmentation}: Openmmlab semantic segmentation toolbox and benchmark. \url{https://github.com/open-mmlab/mmsegmentation} (2020)

\bibitem{RN17}
Hatamizadeh, A., Yang, D., Roth, H.R., Xu, D.: Unetr: Transformers for 3d medical image segmentation. 2022 IEEE/CVF Winter Conference on Applications of Computer Vision (WACV) pp. 1748--1758 (2021)

\bibitem{Lei2024ConDSegAG}
Lei, M., Wu, H., Lv, X., Wang, X.: Condseg: A general medical image segmentation framework via contrast-driven feature enhancement. ArXiv  \textbf{abs/2412.08345} (2024), \url{https://api.semanticscholar.org/CorpusID:274638646}

\bibitem{RN40}
Liao, W., He, J., Luo, X., Wu, M., Shen, Y., Li, C., Xiao, J., Wang, G., Chen, N.: Automatic delineation of gross tumor volume based on magnetic resonance imaging by performing a novel semisupervised learning framework in nasopharyngeal carcinoma. International Journal of Radiation Oncology*Biology*Physics  \textbf{113}(4),  893--902 (2022). \doi{https://doi.org/10.1016/j.ijrobp.2022.03.031}, \url{https://www.sciencedirect.com/science/article/pii/S0360301622002772}

\bibitem{RN18}
Liu, Z., Wang, H., Lei, W., Wang, G.: Csaf-cnn: Cross-layer spatial attention map fusion network for organ-at-risk segmentation in head and neck ct images. In: 2020 IEEE 17th International Symposium on Biomedical Imaging (ISBI). pp. 1522--1525. \doi{10.1109/ISBI45749.2020.9098711}

\bibitem{RN15}
Luo, X., Fu, J., Zhong, Y., Liu, S., Han, B., Astaraki, M., Bendazzoli, S., Toma-Dasu, I., Ye, Y., Chen, Z., Xia, Y., Su, Y., Ye, J., He, J., Xing, Z., Wang, H., Zhu, L., Yang, K., Fang, X., Wang, Z., Lee, C.W., Park, S.J., Chun, J., Ulrich, C., Maier-Hein, K.H., Ndipenoch, N., Miron, A., Li, Y., Zhang, Y., Chen, Y., Bai, L., Huang, J., An, C., Wang, L., Huang, K., Gu, Y., Zhou, T., Zhou, M., Zhang, S., Liao, W., Wang, G., Zhang, S.: Segrap2023: A benchmark of organs-at-risk and gross tumor volume segmentation for radiotherapy planning of nasopharyngeal carcinoma. Medical Image Analysis  \textbf{101},  103447 (2025). \doi{https://doi.org/10.1016/j.media.2024.103447}, \url{https://www.sciencedirect.com/science/article/pii/S1361841524003748}

\bibitem{RN35}
Podobnik, G., Strojan, P., Peterlin, P., Ibragimov, B., Vrtovec, T.: Han-seg: The head and neck organ-at-risk ct and mr segmentation dataset. Medical Physics  \textbf{50}(3),  1917--1927 (2023). \doi{https://doi.org/10.1002/mp.16197}, \url{https://aapm.onlinelibrary.wiley.com/doi/abs/10.1002/mp.16197}

\bibitem{RN2}
Ronneberger, O., Fischer, P., Brox, T.: U-net: Convolutional networks for biomedical image segmentation. ArXiv  \textbf{abs/1505.04597} (2015)

\bibitem{RN27}
Takikawa, T., Acuna, D., Jampani, V., Fidler, S.: Gated-scnn: Gated shape cnns for semantic segmentation. In: 2019 IEEE/CVF International Conference on Computer Vision (ICCV). pp. 5228--5237. \doi{10.1109/ICCV.2019.00533}

\bibitem{RN39}
Tang, H., Chen, X., Liu, Y., Lu, Z., You, J., Yang, M., Yao, S., Zhao, G., Xu, Y., Chen, T., Liu, Y., Xie, X.: Clinically applicable deep learning framework for organs at risk delineation in ct images. Nature Machine Intelligence  \textbf{1}(10),  480--491 (2019). \doi{10.1038/s42256-019-0099-z}, \url{https://doi.org/10.1038/s42256-019-0099-z}

\bibitem{RN13}
Vaswani, A., Shazeer, N.M., Parmar, N., Uszkoreit, J., Jones, L., Gomez, A.N., Kaiser, L., Polosukhin, I.: Attention is all you need. In: Neural Information Processing Systems

\bibitem{Wu2019FastFCNRD}
Wu, H., Zhang, J., Huang, K., Liang, K., Yu, Y.: Fastfcn: Rethinking dilated convolution in the backbone for semantic segmentation. ArXiv  \textbf{abs/1903.11816} (2019), \url{https://api.semanticscholar.org/CorpusID:85542864}

\bibitem{RN14}
Xie, E., Wang, W., Yu, Z., Anandkumar, A., Álvarez, J.M., Luo, P.: Segformer: Simple and efficient design for semantic segmentation with transformers. In: Neural Information Processing Systems

\bibitem{RN22}
Yan, K., Yin, X., Xia, Y., Wang, F., Wang, S., Gao, Y., Yao, J., Li, C., Bai, X., Zhou, J., Zhang, L., Lu, L., Shi, Y.: Liver Tumor Screening and Diagnosis in CT with Pixel-Lesion-Patient Network (2023). \doi{10.48550/arXiv.2307.08268}

\bibitem{RN23}
Yuan, Y., Hou, S., Wu, X., Wang, Y., Sun, Y., Yang, Z., Yin, S., Zhang, F.: Application of deep-learning to the automatic segmentation and classification of lateral lymph nodes on ultrasound images of papillary thyroid carcinoma. Asian journal of surgery  (2024)

\bibitem{Zhang2022AcceleratingDC}
Zhang, G., Luo, Z., Yu, Y., Cui, K., Lu, S.: Accelerating detr convergence via semantic-aligned matching. 2022 IEEE/CVF Conference on Computer Vision and Pattern Recognition (CVPR) pp. 939--948 (2022), \url{https://api.semanticscholar.org/CorpusID:247446935}

\bibitem{RN36}
Zhang, Z., Qi, X., Zhang, B., Wu, B., Le, H., Jeong, B., Liao, Z., Liu, Y., Verjans, J., To, M.S., Hartley, R.: Segreg: Segmenting oars by registering mr images and ct annotations. In: 2024 IEEE International Symposium on Biomedical Imaging (ISBI). pp.~1--5. \doi{10.1109/ISBI56570.2024.10635437}

\bibitem{RN31}
Zhu, X., Su, W., Lu, L., Li, B., Wang, X., Dai, J.: Deformable detr: Deformable transformers for end-to-end object detection. ArXiv  \textbf{abs/2010.04159} (2020)

\end{thebibliography}
%




\end{document}